# Enhancing Depressive Post Detection in Bangla: A Comparative Study of TF-IDF, BERT and FastText Embeddings

**Saad Ahmed Sazan[1], Mahdi H. Miraz[2,3,4,*] and A B M Muntasir Rahman[5]**

[1]Rajshahi University of Engineering and Technology, Bangladesh
1803020@student.ruet.ac.bd
[2]Xiamen University Malaysia, Malaysia
m.miraz@ieee.org
[3]Wrexham University, UK
m.miraz@ieee.org
[4]University of South Wales, UK
m.miraz@ieee.org
[5]Brac University, Bangladesh
abm.muntasir@bracu.ac.bd
*Correspondence: m.miraz@ieee.org

**Abstract:** Due to massive adoption of social media, detection of users' depression through social media analytics bears significant importance, particularly for underrepresented languages, such as Bangla. This study introduces a well-grounded approach to identify depressive social media posts in Bangla, by employing advanced natural language processing techniques. The dataset used in this work, annotated by domain experts, includes both depressive and non-depressive posts, ensuring high-quality data for model training and evaluation. To address the prevalent issue of class imbalance, we utilised random oversampling for the minority class, thereby enhancing the model's ability to accurately detect depressive posts. We explored various numerical representation techniques, including Term Frequency – Inverse Document Frequency (TF-IDF), Bidirectional Encoder Representations from Transformers (BERT) embedding and FastText embedding, by integrating them with a deep learning-based Convolutional Neural Network-Bidirectional Long Short-Term Memory (CNN-BiLSTM) model. The results obtained through extensive experimentation, indicate that the BERT approach performed better the others, achieving a F1-score of 84%. This indicates that BERT, in combination with the CNN-BiLSTM architecture, effectively recognises the nuances of Bangla texts relevant to depressive contents. Comparative analysis with the existing state-of-the-art methods demonstrates that our approach with BERT embedding performs better than others in terms of evaluation metrics and the reliability of dataset annotations. Our research significantly contribution to the development of reliable tools for detecting depressive posts in the Bangla language. By highlighting the efficacy of different embedding techniques and deep learning models, this study paves the way for improved mental health monitoring through social media platforms.

**Keywords:** *BERT; Bi-LSTM; CNN; Depression; FastText; Post Detection; TF-IDF; Text Classification*

---

## 1. Introduction

Depression is a serious condition characterised by worsening of negative emotions [1]. Prolonged sadness or enduring a difficult situation that causes continuous, unbearable sufferings can lead to depression. If left untreated, depression can even lead to the suicide [2]. According to the World Health Organization (WHO), an estimated 280 million people, i.e. 3.8% of the global population, experience depression, including 5% adults and 5.7% senior citizens (60 years or older)[1]. In 2022, 49,369 people worldwide took their own lives due to depressive disorders, and the rate of suicide continues to rise[2].In era of this digitally connected word, social media is everywhere, serving as a platform for people to express themselves, communicate and share information [3]. However, social media often becomes a space where people express mental health issues such as depression [4]. It is important to accurately and promptly identify these depressive posts on social media so that early intervention, timely support and potentially life-saving actions can be taken. Machine learning approaches have emerged as powerful tools to address this challenge. They can process and analyse large amounts of data, uncovering patterns that may not be visible to the human eye. The necessity of depressive post detection on social media stems from the necessity of timely assistance and intervention. Social media platforms provide a unique window into the thoughts and emotions of individuals in real time. By analysing the language used in posts, machine learning models can identify linguistic cues and patterns associated with depression. This automated detection can then trigger alerts for mental health professionals or support systems, ensuring that the individuals promptly receive the help they need. This scalable approach offers a level of objectivity and consistency that is difficult to achieve through manual monitoring on its own.

While there has been noteworthy progress in detecting depressive posts in widely spoken languages, such as English, not much advancement in this domain is apparent for less widely spoken languages, such as Bangla. Bangla, spoken by over 230 million people worldwide[3], mainly in Bangladesh and the Indian state of West Bengal, presents unique linguistic challenges for natural language processing (NLP) [5]. The scarcity of Bangla datasets, linguistic resources and pre-trained models makes it even more difficult to detect depressive posts in this language. Figure 1 demonstrates a significant rise in depressive disorders in Bangladesh over the past three decades[4]. This linear upward trend highlights a critical public health concern, underscoring the importance of identifying individuals suffering from depression and implementing effective interventions. Addressing this growing mental health issue is crucial for improving the well-being and quality of life of the Bangladeshi population.

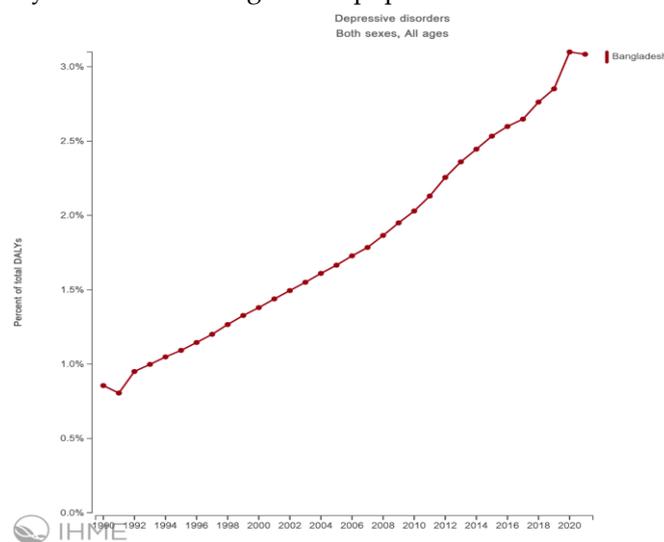

**Figure 1.** Percentage of Depressive Disorder to total DALY's in Bangladesh (1990-2021)[4]

This research aims to enhance depressive post detection in Bangla by exploring and comparing three different embedding techniques: Term Frequency-Inverse Document Frequency (TF-IDF) [6], Bidirectional Encoder Representations from Transformers (BERT) [7] and FastText embeddings [8]. Each technique offers distinct advantages compared to others. TF-IDF is a well-established method that assigns importance to the words based on their frequency in a particular document relative to a corpus; BERT captures deep contextualised embeddings by understanding the nuances of language through a transformer architecture;

---

[1] https://www.who.int/news-room/fact-sheets/detail/depression#:~:text=Approximately%20280%20million%20people%20in
[2] https://www.kff.org/mental-health/issue-brief/a-look-at-the-latest-suicide-data-and-change-over-the-last-decade/
[3] https://bangladeshus.com/roots-of-the-bangla-language/#:~:text=The%20Bangla%20language%2C%20also%20known%20as%20Bengali%2C%20is
[4] https://vizhub.healthdata.org/gbd-compare/

and FastText generates word embeddings that can handle out-of-vocabulary words by utilising Sub-word information.

By comparing the performance of TF-IDF, BERT and FastText embeddings for Bangla depressive post detection, this study seeks to identify the most effective approach. The goal is to enhance the understanding of depressive language in Bangla and pave the way for developing tools that can support mental health monitoring on social media. The key contributions of our work include:

1. **Effectively Handling Class Imbalance:** We addressed the issue of class imbalance in our dataset, which ensures that our model performs better to predict the minority class.
2. **Enhancing Performance of Text Representation Techniques**: Our research demonstrates that while Term Frequency-Inverse Document Frequency (TF-IDF) effectively captures the essential features, BERT embeddings provide a more comprehensive understanding of the texts. This indicates that while simpler methods (e.g. TF-IDF) are useful, advanced embeddings (e.g. BERT) can offer significant benefits in capturing the nuanced semantics of Bangla depressive posts. This finding underscores the importance of choosing the right text representation technique based on the specific requirements as well as constraints of the dataset and the application.
3. **Proposing a Novel Custom CNN-BiLSTM Model:** We introduce a custom model combining Convolutional Neural Networks (CNN) and Bidirectional Long Short-Term Memory (BiLSTM) networks. This hybrid approach captures local patterns and long-term dependencies, resulting in highly accurate predictions of depressive posts. This model represents a significant advancement in the automatic detection of depressive symptoms from the texts.

## 2. Related Works

While there have been numerous studies conducted in widely spoken languages such as English, there is a notable lack of research in Bangla regarding depressive text analysis. This section provides a comprehensive literature review on this topic.

The study by Uddin *et al.* [7] delved into the understanding of depression through Bangla social media data. The researchers utilised advanced models such as Gated Recurrent Unit (GRU) and Long Short-Term Memory (LSTM) Recurrent Neural Networks. The development of a small yet meticulously organised dataset consisting of Bangla tweets is the novelty of their work. The findings of their research demonstrated that adjusting the settings of these models, also known as hyper-parameter tuning, significantly impacted their accuracy. It has also been reported that the GRU models outperformed the LSTM models, particularly with the smaller dataset used. However, this justification was solely based on the achieved accuracy scores, which can sometimes be misleading. In fact, the reduced size of the dataset can result in a less generalisable model. This implies that the model may perform well on the reduced dataset but poorly on new, unseen data due to the lack of diverse training examples. This constitutes a significant research gap in their work.

Another study by Uddin *et al.* [8] deployed a Long Short-Term Memory (LSTM) Deep Recurrent Network for depression analysis on Bangla social media data. The study involved the creation of a small dataset of Bangla tweets, which was then stratified. The paper demonstrated the impact of hyper-parameter tuning on the efficacy of depression analysis on a small Bangla social media dataset. The data was sourced from Twitter, with 5,000 Bangla tweets collected through repeated sampling, allowing for random repetitions of tweets. The 5,000 Bangla tweets were categorised into four groups: depressive (984 tweets), non-depressive (2930 tweets), ambiguous (699 tweets) and incomplete sentences (387 tweets). The initial dataset exhibited an imbalance, with 2,930 non-depressive tweets and only 984 depressive tweets, which could lead to accuracy and overfitting issues. Consequently, 984 non-depressive tweets were chosen to balance the dataset with 984 depressive tweets, excluding ambiguous and incomplete tweets. The study applied the LSTM Deep Recurrent model to analyse Bangla tweets for predicting human depression. The findings revealed that the LSTM model with a size of 128, batch size of 25 with 10 epochs and 5 layers with 20 epochs achieved high depression detection accuracies. This suggests that high accuracy can be achieved for small datasets in complex psychological tasks such as depression analysis by tuning the Deep Recurrent model. However, it is important to note that this approach has limitations similar to those of the previous work [7]. The justifications provided were solely based on the accuracy metric and down-sampling the dataset may lead to a loss of model generalisation. Moreover, smaller datasets can also impact the reliability of evaluation metrics.

Chowdhury *et al.* [9] aimed to automatically extract sentiment or polarity expressed by users in Bangla Twitter posts or "tweets". Since no labelled training corpus of Bangla tweets was available, they utilised tweets obtained from the Twitter API, which were designated for the training set, to construct the dataset for training the classifier. For this purpose, they utilised a semi-supervised method called self-training bootstrapping. Their practical findings were promising for a resource-scarce language such as Bangla, achieving an accuracy of 93% for SVM, using unigrams with emoticons as features. The underlying assumption was that users share tweets to convey opinions and subjective content, effectively narrowing down the classification problem to identifying the overall polarity of tweets as either negative or positive. To construct the training corpus, they employed a semi-supervised bootstrapping approach, eliminating the need for labour-intensive manual annotation. Drawing from previous research in English, Support Vector Machine (SVM) and Maximum Entropy (MaxEnt) were found to outperform other classifiers in this domain. Consequently, for classification, they utilised SVM and MaxEnt, conducting a comparative analysis of the performance of these two machine learning algorithms by experimenting with various sets of features. However, it is worth noting that their dataset was labelled by the people who are not expert in identifying the polarity which is one of the key limitations of their work. Non-experts could introduce biases based on their subjective understanding of the data, potentially skewing the model's performance and creating a bias toward non-expert perspectives.

Tasnim *et al.* [10] employed various machine learning algorithms to detect depressive Bangla text from social media posts. Feature extraction methods such as count vectorisation, TF-IDF and word embedding were applied to a dataset containing 6,178 texts gathered from social media. The dataset was self-generated and perfectly balanced. The experiment utilised Multinomial NB, Aggressive Classifier, Decision tree classifier, Neural Network and Linear Support Vector Machine. Additionally, two deep learning models, namely Bidirectional LSTM (BiLSTM) and Gated Recurrent Units (GRU), were employed. Each model was subjected to 10-fold cross-validation, with each fold serving as a test dataset for the corresponding training iteration. Their research achieved a classification accuracy of 97% using the decision tree algorithm and 94% with the bidirectional LSTM deep learning model for predicting depressive text in the Bangla language. However, similar to the work by Chowdhury *et al.* [9], this study also faced limitations related to the labelling of the dataset by non-experts.

Akhter *et al.* [11] proposed the use of machine learning algorithms and user information to detect cyberbullying in Bangla text. They gathered a dataset from social media, labelling it as bullied or not bullied, to train various classification models. Cross-validation results showed that a support vector machine (SVM) algorithm achieved a 97% detection accuracy. The study aimed to develop a novel method for analysing Bangla content on social media by combining text analytics and machine learning algorithms and to compare its performance with other techniques. They extensively explored suitable algorithms for Bangla text categorisation, including Naive Bayes, SVM, Decision Tree and K-Nearest Neighbours, using the WEKA software platform. The experiments involved 2,400 Bangla texts from social media posts, with 10% labelled as bullying, and a 10-fold cross-validation model was used to evaluate the models' performance. It is worth noting that the model was trained using an imbalanced dataset, which can lead to bias towards the majority class. Additionally, similar to [9] and [10], the dataset was not labelled by an expert, potentially leading to biased classification.

In the work by Hassan *et al.* [12], a substantial textual dataset of both Bangla and Romanised Bangla texts was provided, marking the first of its kind. This dataset underwent pre-processing and multiple validations and was made ready for sentiment analysis (SA) implementation and experiments. Furthermore, the dataset was tested in a Deep Recurrent model, specifically Long Short-Term Memory (LSTM), using two types of loss functions: binary cross-entropy and categorical cross-entropy. Experimental pre-training was also conducted by utilising data from one validation set to pre-train the other and vice versa. Their key contributions included providing a dataset comprising 10,000 Bangla and Romanised Bangla text samples, each annotated by two adult Bangla speakers, pre-processing the data for easy usability by researchers, applying deep recurrent models to the Bangla and Romanised Bangla text corpus and experimenting with pre-training the dataset of one label for another (and vice versa) to assess potential improvements in results. It is important to consider that the achieved level of accuracy may not be deemed fair. Moreover, solely prioritising accuracy without considering other factors may raise concerns about the validity of the justification.

These studies exhibit several common limitations, including the absence of datasets labelled by experts, the training of models on imbalanced datasets, and the reliance solely on accuracy scores, which can be misleading. Our research primarily addresses these limitations. We utilised a dataset labelled by domain experts, addressed the issue of class imbalance and provided a comprehensive set of evaluation metrics to offer a holistic view of our model's performance.

## 3. Methodology and System Architecture

Our research introduces a comprehensive method to predict depressive Bangla posts, aiming to accurately determine whether a social media post exhibits any depressive characteristics or not. The approach begins with dividing the dataset into training and testing sets to facilitate model validation. To address the imbalance in the minority class, which often poses a challenge in such datasets, random oversampling techniques have been employed. This step ensures that the training set has a more balanced representation of depressive and non-depressive posts. Next, the text undergoes preprocessing to remove noise, such as irrelevant characters and symbols, enhancing the quality of the input data. Word embedding for text vectorisation has then been utilised, transforming the textual data into numerical vectors which can be effectively processed by machine learning algorithms. With the pre-processed and vectorised data, a deep learning model tailored to predict depressive Bangla posts has then been constructed and trained. This model leverages the capabilities of neural networks to capture complex patterns and nuances in the text, improving prediction accuracy. An overview of our proposed approach, including all these steps, is illustrated in Figure 2, providing a visual representation of the entire process from data preparation to model training and evaluation. Algorithm 1 illustrates the pseudocode of the proposed methodology.

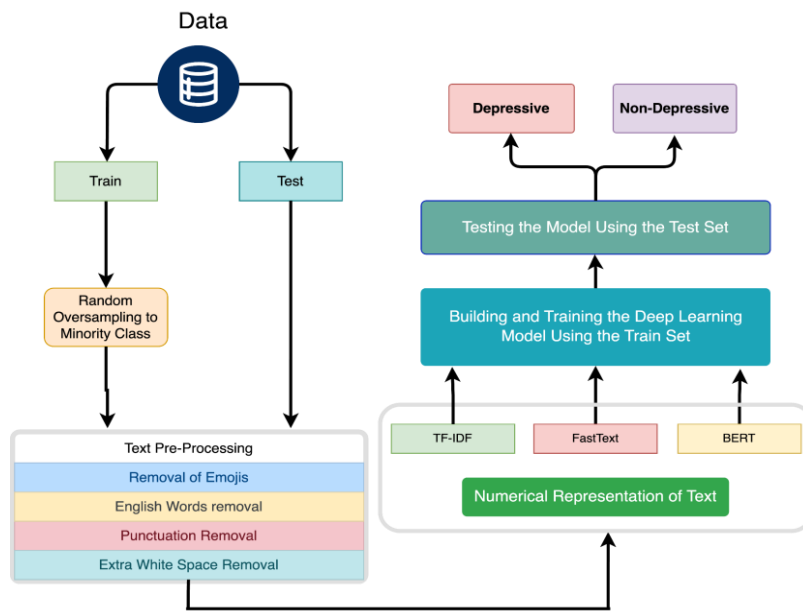

**Figure 2.** Flow diagram of the Proposed Methodology

**Algorithm 1.** Proposed methodology

1. Data Preparation
    1.1 Import necessary libraries
    1.2 Load Data
        1.2.1 Load the dataset containing Bangla depressive and non-depressive post
        1.2.2 Split the data into text (X) and label (y)
    1.3 Preprocess Data
        1.3.1 Tokenize and clean the text data
        1.3.2 Apply TF-IDF/BERT/FastText vectorization to convert text data into numerical vectors.
    1.4 Handle class imbalance using random over sampler method
2. Model Training
    2.1 Build the CNN-BiLSTM Model

        2.1.1     Initialize a sequential model
        2.1.2     Add Convolutional layers for feature extraction
        2.1.3     Add MaxPooling layer to down-sample the extracted features
        2.1.4     Add Bidirectional LSTM layers to capture contextual information
        2.1.5     Add Dense layers for the final classification
    2.2     Compile the model
    2.3     Train the model
        2.3.1     Fit the model to training data with a specified number of epochs and batch size
        2.3.2     Use validation set to monitor the training progress
3.     Evaluate the Model

### 3.1. Dataset

This segment provides a description of the dataset. The dataset was created by Uddin *et al.* [7] and available on GitHub[5]. Initially, they collected the data from various tweets. As a secondary source, they also collected Bangla depressive data using user through distributing online questionnaire. The dataset was manually labelled by a sociology student, as a domain expert in human behaviours. There are a total of 2930 non-depressive posts and 984 depressive posts in the dataset. This dataset is the largest publicly available collection of Bangla depressive and non-depressive posts. Additionally, this dataset has been meticulously labelled by experts, ensuring high-quality and reliable annotations, which is a key factor in our selection. While there are other datasets, they are not publicly accessible and most of them lack expert annotation. These factors make the selected dataset[5] the optimal choice for our study. While the class distribution of the dataset is shown in Figure 3, the sample of the dataset is depicted in Table 1, showing two different posts from two distinct classes.

**Table 1.** Snapshot of the Dataset

| Text | Label |
|---|---|
| স্বার্থক জন্ম আমরা জন্মেছি এই দেশে রাস্তায় মানুষ মরে আর মন্ত্রী সাহেব হাসে হায় আফসোস | Depressive |
| জিত্তা গেছি ! বাহ্ ! এটাই বাংলাদেশ ! সাবাস ! অভিনন্দন ! ২০০৯সালের পর এশিয়ার বাইরে এটাই ওডিআই সিরিজ জয় ! | Non_depressive |

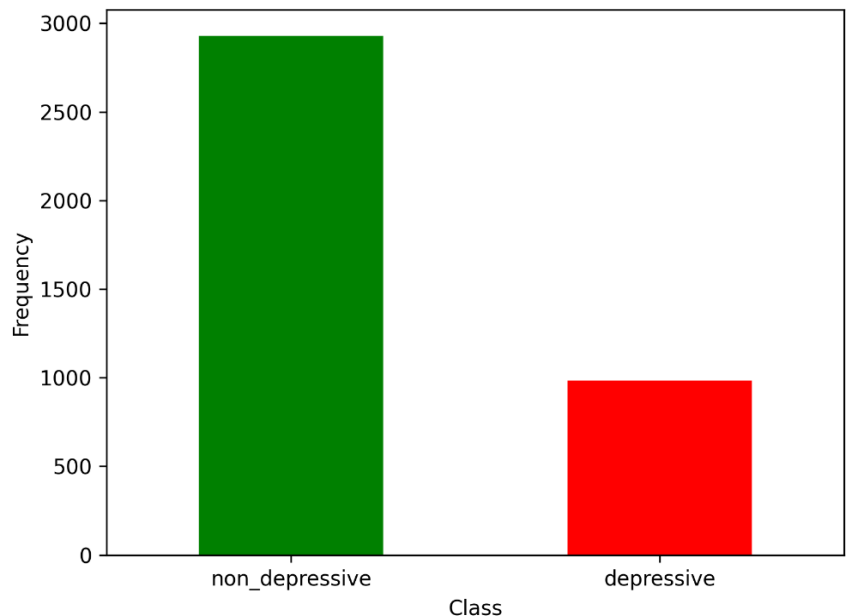

**Figure 3.** Class Distribution

It is apparent that the dataset is significantly imbalanced. The number of non-depressive posts is much higher than that of depressive posts. Figure 4 shows the most frequent 50 words in depressive and non-depressive posts in different word clouds. Figure 5 illustrates the length distribution of depressive and non-depressive posts. It is evident that non-depressive posts are more verbose than depressive ones, indicating that people are more concise when writing depressive posts.

---

[5] https://github.com/abdulhasibuddin/Depression-Analysis-from-Social-Media-Data-in-Bangla-Language-Applying-Deep-Recurrent-Neural-Network/tree/master/Implementations

### 3.2. Data Split and Random Oversampling

As the dataset is imbalanced, training a model with it will be biased, and the evaluation metrics will not reflect the appropriate results. We have incorporated random oversampling to deal with this issue. Random oversampling is a technique used to address class imbalance in datasets where one class is significantly underrepresented compared to the others. Random oversampling increases the number of instances of the minority class by randomly duplicating them.

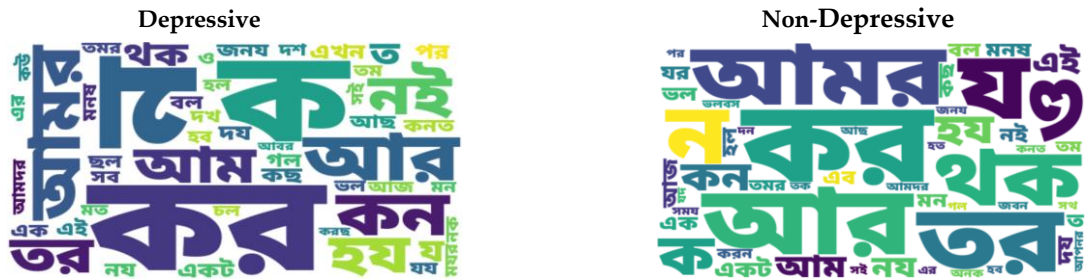

**Figure 4.** Word Clouds for Depressive and Non-Depressive Post

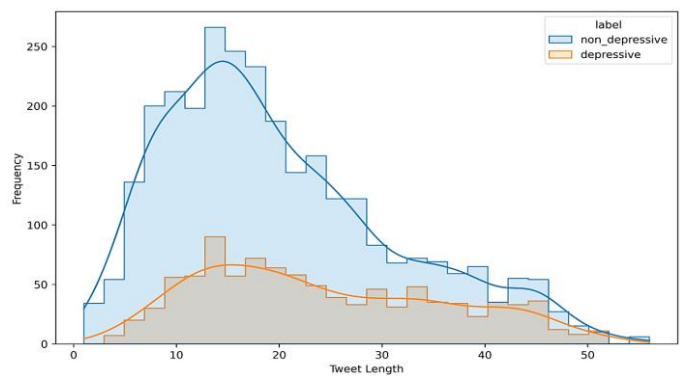

**Figure 5.** Post Length Distribution

This results in a more balanced class distribution, which helps the model learning to more effectively recognise the patterns for the both classes.

It is important to note that there is a necessity to oversample the training set instead of the whole dataset, because oversampling the entire dataset could lead to data leakage when performing cross-validation or evaluating the model on a separate test set. Data leakage occurs when information from the test set leaks into the training process, leading to overly optimistic performance estimates. By oversampling only the training set, it has been ensured that the test set remains untouched and representative of real-world data. The data split was done in a ratio of 70-30, i.e. 70% of the data was kept for training and the remaining 30% for testing. Therefore, amongst the 3914 posts, 2739 were initially kept for training and the remaining 1175–were kept for testing. Then, 20% of the training data was kept for validation purposes during the training process.

### 3.3. Text Pre-processing

Text preprocessing is an essential step in natural language processing (NLP) that involves cleaning and transforming raw text into a format that is suitable for analysis and modelling. Proper preprocessing can significantly improve the performance of NLP models by reducing noise and ensuring consistency in the data. In our dataset, there are some noises that should be pre-processed before being passed to the model for analysis. For an optimal result, the following preprocessing steps have been conducted:

1. **Removing Emojis:** Emojis can introduce noise into the dataset. Removing emojis ensures that the text is standardised and focuses on the language content, making it easier to process and analyse.
2. **Removing English Words:** Our primary goal is to analyse Bangla text; English words might not be relevant and can distort the analysis. Removing them helps in focusing the analysis strictly on Bangla language patterns.

3. **Removing Punctuations:** Punctuation marks can sometimes interfere with the text processing algorithms, especially those which are not punctuation-aware. Removing them can simplify the text, making it easier for the models to learn patterns in the data.
4. **Removing Extra White Spaces:** Extra white spaces can create inconsistencies in the dataset. They can affect tokenisation and other preprocessing steps, leading to irregularities in text representation. Removing unnecessary white spaces ensures that the text data is clean and compact, making it more efficient for storage and faster for processing.

For instance, the sentence 'শুভ সকাল পবিত্র জুম্মার দিন জুম্মা মোবারক ! 🌹সুন্দর হোক সবার জীবন #Jumma' will be transformed to 'শুভ সকাল পবিত্র জুম্মার দিন জুম্মা মোবারক সুন্দর হোক সবার জীবন' after the preprocessing steps.

### 3.4. Numerical Representation of Text

Numerical representation of text refers to the process of transforming text data into numerical vectors or matrices that can be processed by machine learning algorithms or other mathematical models. This conversion is necessary because most machine learning algorithms require numerical input data to perform computations and effectively learn patterns. In our work, different representations of textual data, including TF-IDF, BERT embedding and FastText embedding, have been employed.

#### 3.4.1. TF-IDF

Term Frequency-Inverse Document Frequency (TF-IDF) is a statistical measure used in information retrieval and text mining to evaluate the importance of a word in a document relative to a collection of documents (corpus). It is a numerical statistic intended to reflect how important a word is to a document in a corpus. The importance proportionally increases to the number of times a word appears in the document but is offset by the frequency of the word in the corpus, which helps to adjust for the fact that some words appear more frequently in general. TF measures how frequently a term appears in a document, while IDF measures how important a term is in the entire corpus. The formulas for finding TF and IDF are as follows:

$$TF\ (t,d) = \frac{Number\ of\ times\ term\ t\ appears\ in\ document\ d}{Total\ number\ of\ terms\ in\ document\ d} \tag{1}$$

$$IDF\ (t) = log(\frac{Total\ number\ of\ documents}{Number\ of\ documents\ containing\ term\ t}) \tag{2}$$

$$TF - IDF\ (t,d) = TF(t,d) \times IDF(t) \tag{3}$$

#### 3.4.2. Word Embedding

Word embeddings are a type of word representation that allows words to be mapped to the vectors of real numbers. Each word is represented as a dense vector with fixed dimensions. These vectors capture the semantic meanings of the words in a continuous vector space, where words with similar meanings are located close to each other. BERT and FastText are both powerful word-embedding models used in natural language processing tasks. While BERT provides deep contextualised embeddings that capture bidirectional context, FastText leverages Sub-word information to effectively handle morphological variations.

- **BERT Embedding:** BERT is a state-of-the-art pre-trained language representation model developed by Google [7]. It uses a transformer-based architecture, which enables it to capture bidirectional contexts in the texts. BERT is pre-trained on large text corpora using masked language modelling (MLM) and next-sentence prediction (NSP) tasks. BERT generates embeddings by tokenising input text, mapping tokens to initial word embeddings and passing them through multiple transformer encoder layers to capture contextualised representations. These embeddings provide a deep contextual understanding of words within sentences, allowing for accurate processing of natural language data across various tasks.
- **FastText Embedding:** FastText is a word embedding model developed by Facebook's AI Research (FAIR) lab [8]. FastText embedding works by leveraging sub-word information, breaking down words into smaller units such as character n-Grams and training word embeddings using a skip-gram model with negative sampling. During the training process, the model learns to predict the context (surrounding words) given a target word based on its sub-word representations, adjusting parameters to minimise the loss. The resulting embeddings provide dense vector representations for words, capturing semantic relationships and meanings based on their sub-word compositions.

### 3.5. Developing Deep Learning-based (CNN-BiLSTM) Model

The CNN-BiLSTM model is a sophisticated neural network architecture that combines Convolutional Neural Networks (CNNs) [13] and Bidirectional Long Short-Term Memory (BiLSTM) [14] networks which can efficiently predict binary classes in sequential data. The model begins with a Conv1D layer that applies 100 filters of size 3 to the input data, effectively capturing local patterns through the use of the ReLU [15] activation function, which introduces non-linearity. This is followed by a BatchNormalisation [16] layer that stabilises the training process by maintaining the mean and variance of the activations close to 0 and 1, respectively. A MaxPooling1D [17] layer then reduces the dimensionality of the feature maps, retaining the most significant features while decreasing computational load.

The core of the model's ability to handle sequential data lies in the Bidirectional LSTM layer, which processes the input sequence in both forward and backward directions. This bidirectional approach captures dependencies from both the past and the future states, providing a comprehensive understanding of the context. Another BatchNormalisation layer follows, further estabilising and normalising the outputs from the BiLSTM layer.

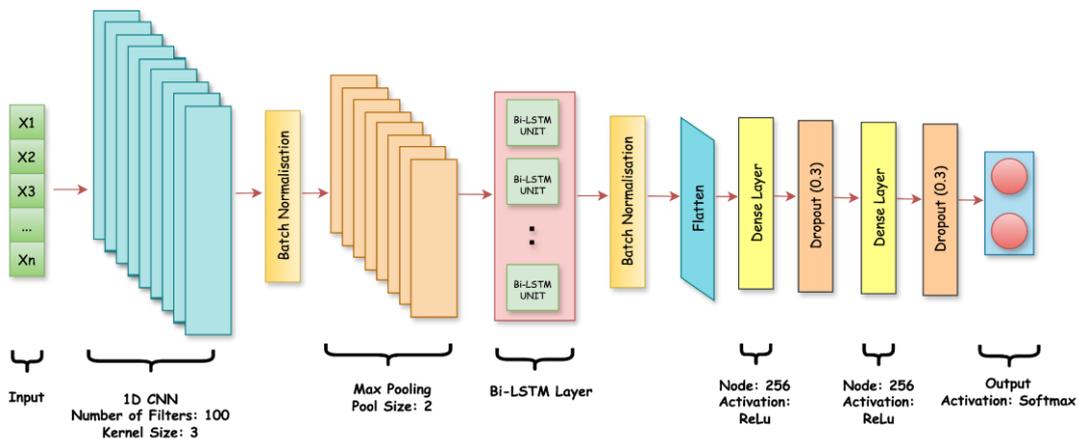

**Figure 6.** Proposed System Architecture (CNN-BiLSTM)

The subsequent flatten layer converts the 2D matrix output into a 1D vector, preparing it for the fully connected dense layers. The first dense layer, with 256 neurons, employs the ReLU activation function and l2 regularisation to learn complex representations of the input features, while preventing overfitting. A Dropout layer [18] with a 30% dropout rate adds further regularisation by randomly dropping neurons during training. This process is repeated with a second dense layer containing 128 neurons and a higher l2 regularisation value, followed by another Dropout layer.

The final output layer consists of two neurons and uses the softmax activation function to output probabilities for each class, facilitating binary classification. The use of softmax allows the model to clearly distinguish between the two classes based on the computed probabilities. The architecture of our proposed CNN-BiLSTM model is shown in Figure 6. The total number of parameters required for the CNN-BiLSTM architecture is 10,034,594, with 10,033,882 trainable parameters, as illustrated in Table 2.

Overall, the combination of CNN for feature extraction, BiLSTM for contextual understanding and various normalisation and regularisation techniques ensures that the model effectively trains, performs effectively on new, unobserved data and efficiently predicts binary classes in sequential datasets.

#### 3.5.1. Model Compilation

The model was compiled using the Adam optimiser [19], which has a learning rate of 0.001 (the default) and categorical_crossentropy as its loss function. Adam optimiser has been chosen because of its adaptive learning rates, which combine the benefits of momentum and RMSprop optimisers as well as require less manual hyperparameter tuning.

### 3.5.2. Model Training

The model was trained using 60 epochs and a batch size of 16. For validation, the training set was split into an 80-20 ratio, with 20% of the data used for validation during the training process. To prevent overfitting, early stopping and an adjustable learning rate were incorporated as callbacks. Early stopping monitored the validation loss with a patience level set to 10. The ReduceLROnPlateau callback continuously monitored the validation loss and reduced the learning rate by a factor of 0.2 if the validation loss did not improve for 2 epochs. This reduction helped fine-tune the model by allowing it to take smaller steps in the parameter space, particularly when the model was close to converging but needed finer adjustments for optimal performance. The entire experiment was conducted on a Kaggle notebook using a CPU accelerator.

**Table 2.** Internal Parameters of the Proposed Model

| Layer (Type) | Output Shape | Parameters |
|---|---|---|
| Conv1D | (None, 298, 100) | 400 |
| BatchNormalization | (None, 298, 100) | 400 |
| MaxPooling1D | (None, 149, 100) | 0 |
| Bidirectional | (None, 149, 256) | 234,496 |
| BatchNormalization | (None, 149, 256) | 1,024 |
| Flatten | (None, 38144) | 0 |
| Dense | (None, 256) | 9,765,120 |
| Dropout | (None, 256) | 0 |
| Dense | (None, 128) | 32,896 |
| Dropout | (None, 128) | 0 |
| Dense | (None, 2) | 258 |

## 4. Experimental Results and Performance Analysis

As part of our research, several experiments were conducted to evaluate the performance of our proposed CNN-BiLSTM model using different textual representation techniques. Our objective was to determine and compare the model's performance using these techniques. Figure 7 and Figure 8 show the accuracy and loss curves of the proposed CNN-BiLSTM model when TF-IDF has been used for the vector representation of the textual data. Figure 8 shows that the model continuously learned from the mistakes, as the loss decreased for the case of TF-IDF. Figure 7 shows the model's performance on the basis of the accuracy of the validation data along with the training data. Figure 9 depicts the confusion matrix. The number of false positives and false negatives is negligible compared to that of the true positives and false positives. Figure 10 shows the ROC (Receiver Operating Characteristic) curve with an AUC (Area Under the Curve) [20] of 84%, exhibiting the model's high ability to distinguish between positive and negative classes.

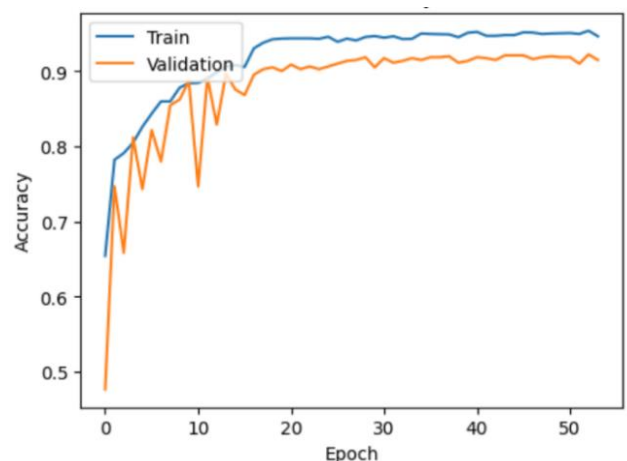
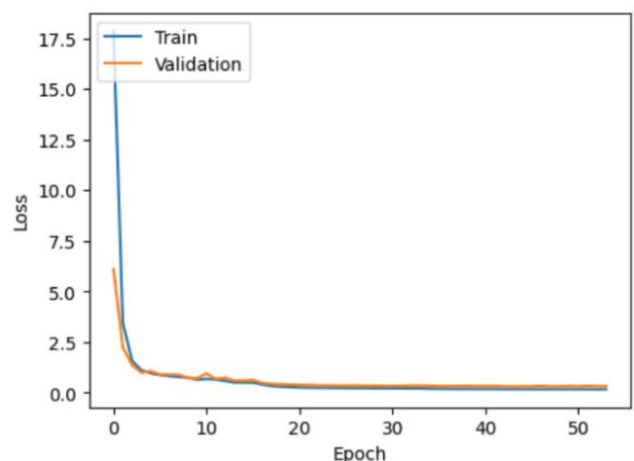

**Figure 7.** Accuracy Curve (TF-IDF)    **Figure 8.** Loss Curve (TF-IDF)

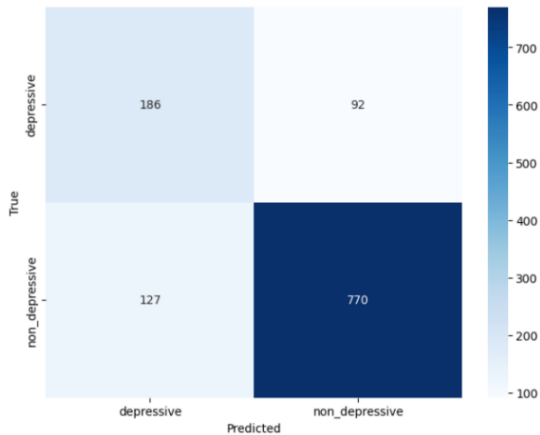
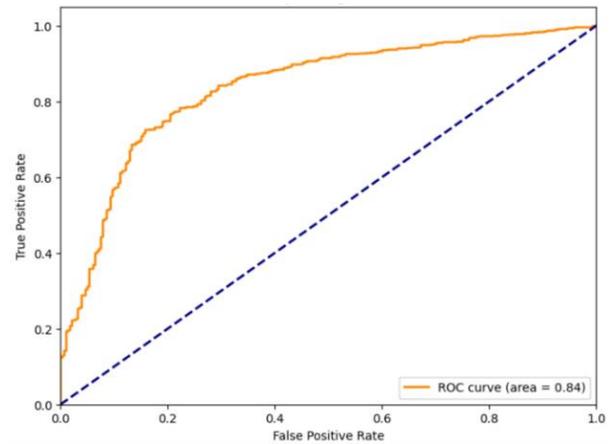

**Figure 9.** Confusion Matrix (TF-IDF)  **Figure 10.** ROC Curve (TF-IDF)

In the case of BERT embedding, the model shows a slight deterioration in the model's generalisation performance compared to the TF-IDF approach. Figure 11 and 12 show the accuracy and loss curve in this case. Though the loss continuously decreased, the model shows a noticeable variance in accuracy for the training and validation set. The reason for this kind of behaviour can be understood by observing when TF-IDF performs better and when BERT performs better. TF-IDF may effectively capture essential information when working with smaller datasets. BERT embeddings, on the other hand, capture more complex semantic relationships in the texts but may require a larger dataset to effectively learn these representations. Figure 13 and Figure 14 show the confusion matrix and the ROC curve for this approach, respectively. The model shows decent performance on the unseen test data, as the ROC covers 81% of the area.

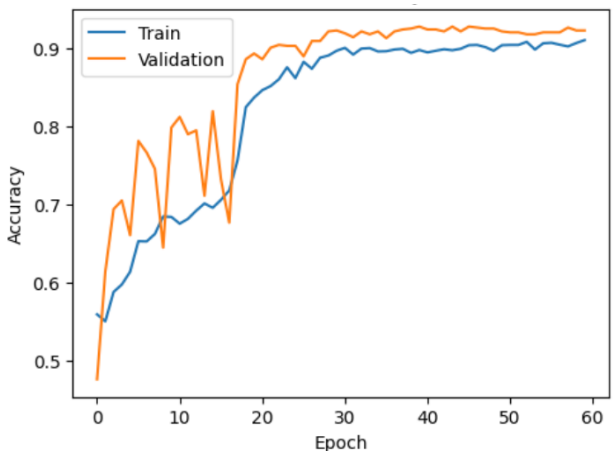
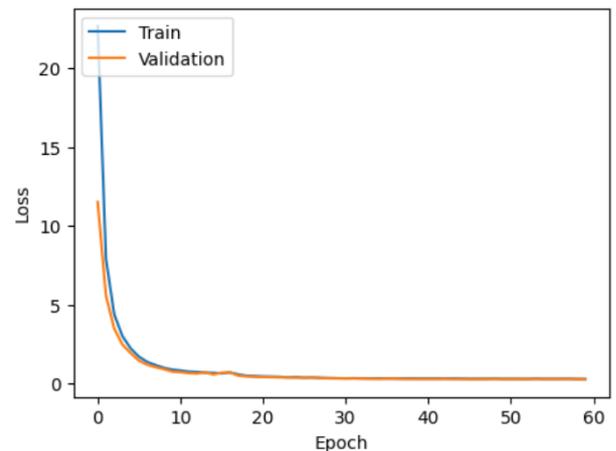

**Figure 11.** Accuracy Curve (BERT Embedding)  **Figure 12.** Loss Curve (BERT Embedding)

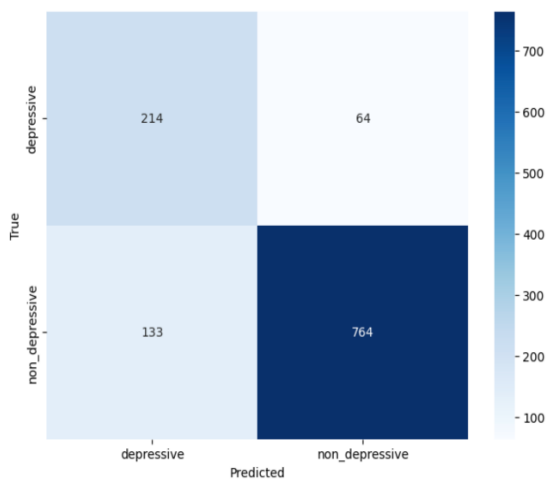
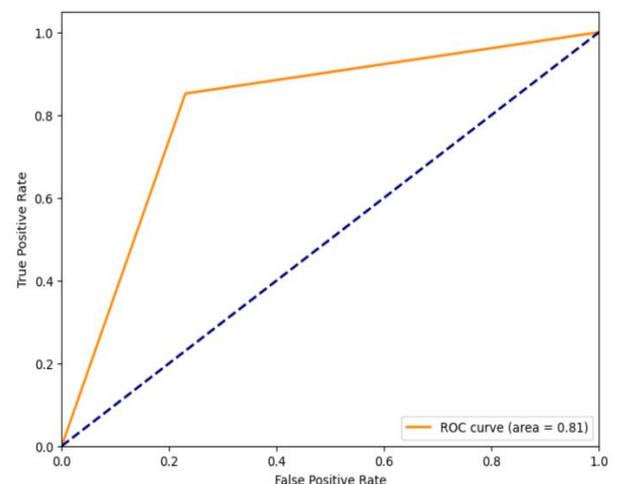

**Figure 13.** Confusion Matrix (BERT Embedding)  **Figure 14.** ROC Curve (BERT Embedding)

The model's performance using BERT and FastText embeddings is almost identical, as shown in Figure 15, where the validation set's accuracy shows minimal improvement over time despite a continuous decrease in loss (Figure 16), indicating slight overfitting. Figures 17 and 18 display the confusion matrix and ROC curve for the FastText approach. The ROC covers 80% of the area, which closely resembles that of the BERT approach. This similarity is attributed to the dataset's characteristics; a small or limited vocabulary dataset prevents FastText from fully leveraging its advantage of capturing sub-word information and producing out-of-vocabulary word embeddings. Consequently, FastText's potential benefits are not fully realised, resulting in comparable performance to BERT, which excels at handling larger, more diverse datasets. Thus, in scenarios with constrained datasets, the choice between BERT and FastText embeddings might not significantly impact the model's performance.

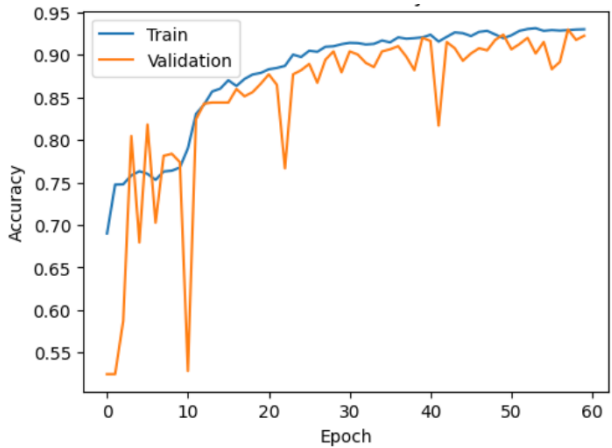

**Figure 15.** Accuracy Curve (FastText Embedding)

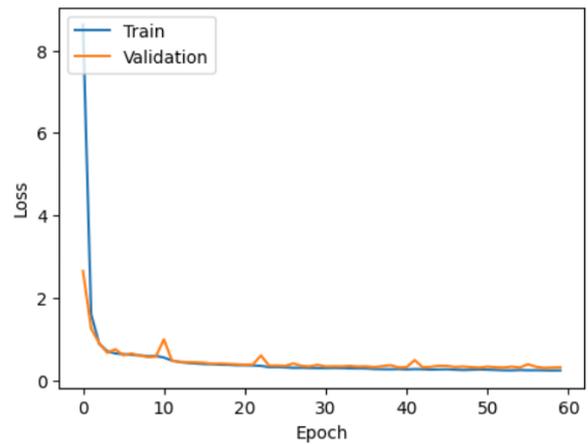

**Figure 16.** Loss Curve (FastText Embedding)

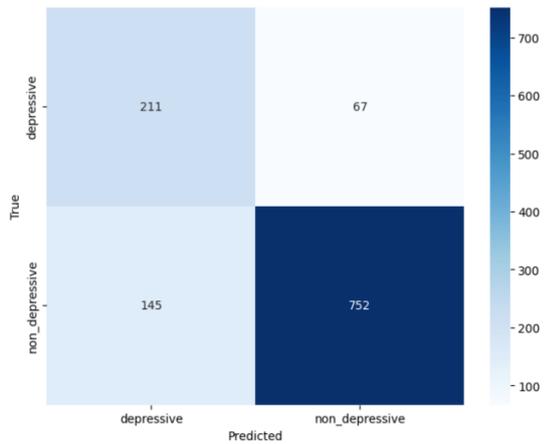

**Figure 17.** Confusion Matrix (FastText Embedding)

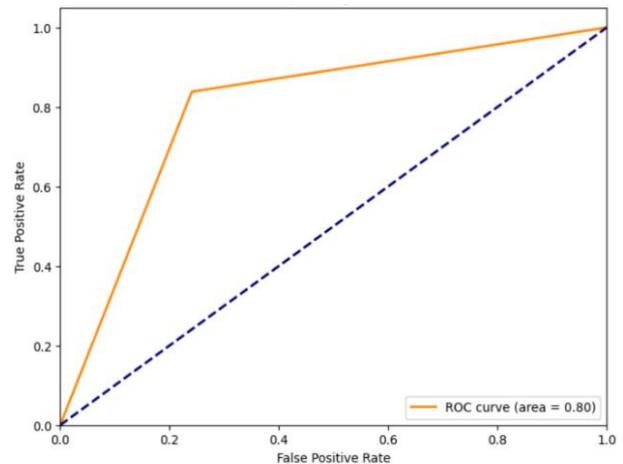

**Figure 18.** ROC Curve (FastText Embedding)

In Table 3, the performance of three different word representation approaches across four evaluation metrics is presented. The BERT representation achieved the highest F1 score of 84%. In comparison, TF-IDF and FastText embeddings achieved slightly lower scores of 82% and 83%, respectively. Moving on to accuracy, TF-IDF, BERT and FastText achieved scores of 81%, 83%, and 82%, respectively. In terms of precision and recall, similar patterns emerged. While the precision scores are 82%, 85% and 84%, respectively, the recall scores are similar to the accuracy scores. These results indicate that the BERT approach achieves the highest F1 score, while FastText and TF-IDF displayed comparable performance in accuracy, precision and recall. Despite TF-IDF achieving the highest AUC (84%), its F1 score (82%), accuracy (81%), and precision (82%) are slightly lower compared to BERT and FastText. This discrepancy implies that while TF-IDF is overall good at distinguishing between classes, it might not be as effective as BERT in handling the balance between precision and recall.

BERT embeddings emerge as the most effective technique for detecting depressive posts, as evidenced by the highest F1 score. This makes it a preferable choice when the goal is to achieve a balanced model with both high precision and recall. While TF-IDF provides a strong AUC score, suggesting good overall discrimination ability, its lower F1 score highlights potential issues in balancing false positives and negatives. In fact, FastText offers a middle ground with consistent performance, making it a reliable alternative.

Table 3. Evaluation Metrics for Three Different Approaches

| Type of Numerical Representation | Accuracy (%) | Precision (%) | Recall (%) | F1-Score (%) |
|---|---|---|---|---|
| TF-IDF | 81 | 82 | 81 | 82 |
| BERT Embedding | 83 | 85 | 83 | 84 |
| FastText Embedding | 82 | 84 | 82 | 83 |

Table 4 provides a comparative overview of recent studies focused on Bangla textual data. Uddin *et al.* [9-10] conducted two distinct studies using the same dataset but employing different methodologies. In the first study, they implemented a GRU model, which yielded an accuracy of 75.7%. However, this study fell short of providing comprehensive evaluation metrics, making it difficult to thoroughly assess the model's performance. The second study used an LSTM model and achieved a higher accuracy of 86.3%, yet it similarly lacked additional performance metrics, which can make relying solely on accuracy potentially misleading. Chowdhury *et al.* [11] achieved an impressive accuracy and F1 score of 93% using a rule-based classifier. However, a notable limitation of their study is that the dataset was manually labelled by the researchers themselves rather than by domain experts, which might introduce biases in the results. In a different approach, Tasnim *et al.* [12] developed a balanced dataset, leading to robust performance metrics across the board. They reported an accuracy of 97%, with precision, recall and F1 measures being consistently high. Despite the high scores achieved, their dataset was also not annotated by experts, as the case of [9-11]. Akhter *et al.* [13] focused on identifying bullying texts and achieved an accuracy of 97.27% using an SVM classifier. However, their dataset was imbalanced, which could affect the generalisability of their findings. Hassan *et al.* [14], on the other hand, used an LSTM model and reported an accuracy of 78% but did not provide any additional evaluation metrics. In contrast to these studies, our research addresses the class imbalance issue within the dataset and employs a BERT embedding combined with a CNN-BiLSTM architecture. This approach resulted in an accuracy of 83%. Unlike several studies that only report accuracy, our study provides a comprehensive set of evaluation metrics, including AUC, precision, recall and F1 score. This holistic evaluation allows for a better understanding of the model's performance beyond mere accuracy, highlighting its ability to effectively balance false positives and false negatives. Furthermore, our dataset was annotated by domain experts, which significantly enhances the reliability and validity of our results. This comprehensive evaluation of both the dataset and the model's performance underscores the robustness of our findings.

Table 4. Comparison with the other published works

| Work | Methodology | Dataset | Accuracy (%) | Precision (%) | Recall (%) | F1 score (%) | Argument |
|---|---|---|---|---|---|---|---|
| Udiin *et al.* [9] | GRU | 984 depressive tweets and 2,930 non-depressive tweets. | 75.7 | - | - | - | Accuracy is fairly low and does not provide other evaluation metrics. |
| Udiin *et al.* [10] | LSTM | 984 depressive tweets and 2,930 non-depressive tweets. | 86.3 | - | - | - | Only accuracy was provided, which might have been misleading. |
| Chowdhury *et al.* [11] | SVM & MaxEnt | Consisted of 1300 unlabelled tweets | 93 | - | - | 93 | Dataset was not labelled by domain experts. |
| Tasnim *et al.* [12] | Decision Tree | 7000 data were collected from Facebook, where 3500 were labelled as depressed and the rest were not depressive. | 97 | 97 | 97 | 97 | Dataset was not labelled by domain experts. |
| Akhter *et al.* [13] | SVM | 2400 texts were collected from Twitter and Facebook where 10% of the data were labelled as bullied. | 97.27 | 99 | - | 99 | The dataset is imbalanced and was not labelled by domain experts. |

| Hassan *et al.* [14] | LSTM | 9337 textual data, where 72% are Romanised Bangla text and the rest are Bangla text. | 78% | - | - | - | The accuracy is not fair enough to make it applicable for real world usage and the dataset was not labelled by domain experts. |
|---|---|---|---|---|---|---|---|
| **Our work** | CNN-BiLSTM | 984 depressive tweets and 2,930 non-depressive tweets. | 83 | 85 | 83 | 84 | The dataset used was labelled by domain experts, and we achieved higher accuracy along other evaluation metrics. This indicates the potential suitability of our model for real-world applications. |

**4.1. Practical Consequences of Findings**

The practical consequences of our findings are significant for several stakeholders, including mental health professionals, researchers and technology developers working with Bangla textual data. Firstly, the ability of our model to accurately identify depressive posts with a balanced approach ensures that the mental health professionals can rely on automated systems to flag potentially at-risk individuals. This early detection is crucial for timely intervention and support, including saving lives. Secondly, the use of BERT embeddings in our CNN-BiLSTM architecture demonstrates the effectiveness of advanced natural language processing techniques in capturing the nuanced semantics of Bangla text. This encourages further research into applying these techniques to other low-resource languages, globally broadening the scope of mental health monitoring tools. Moreover, our comprehensive evaluation metrics, including precision, recall and F1 scores, provide a more transparent and robust assessment of model performance, moving beyond mere accuracy. This transparency is critical for building trust amongst the users and the other stakeholders, ensuring that the model's predictions are reliable and actionable. Lastly, by addressing class imbalance in our dataset, precedent is set for future studies to consider data representativeness, thereby improving the generalisability and fairness of AI systems in mental health applications. Overall, our findings highlight the potential of combining advanced NLP techniques with thoughtful data handling practices to create effective and trustworthy tools for mental health monitoring in Bangla and other underrepresented languages.

**5. Limitations and Future Works**

This section demonstrates the limitations faced throughout this work, along with our future research directions.

**5.1. Limitations**

- **Small Dataset Size**: One key limitation is the small size of our dataset, which restricts the model's ability to generalise to a broader range of data. This constraint limits the scope and applicability of our findings.
- **Test Set Size**: The test set size was not sufficient to conclusively determine the model's performance across a diverse range of inputs. A larger test set would have provided a more robust evaluation of the model's capabilities.
- **Lack of External Validation**: We were unable to test the robustness of our model with other datasets, as they were not publicly available. This limits our ability to confirm the model's performance on different data sources.

**5.2. Future Works**

In the future, we aim to address the above-mentioned limitations by developing a larger public dataset to enhance the robustness and generalisability of our model. Expanding the dataset will allow for more comprehensive testing and validation, ultimately leading to a more reliable tool for detecting depressive posts in Bangla. This larger dataset will facilitate improved model training, ensuring better performance across diverse data samples. Additionally, our plan also includes exploring more advanced techniques and architectures to further improve the detection accuracy and efficiency of our model.

**6. Conclusion**

Our work aimed to contribute to the identification of individuals experiencing ongoing depression, as reflected in their social media posts. Given the scarcity of research in the Bangla language, it is essential to develop a reliable tool for detecting depressive posts written in Bangla. To this end, our research utilised a dataset containing both depressive and non-depressive posts that were carefully annotated by domain experts. To address the issue of class imbalance, random oversampling was employed for the minority class, ensuring that our model could effectively learn from a more balanced dataset. This step was crucial in preventing the model from being biased towards the majority class and enhancing its ability to accurately detect depressive posts. Several experiments were conducted with various numerical representation techniques to convert textual data into meaningful vectors for machine learning. These techniques included Term Frequency-Inverse Document Frequency (TF-IDF), BERT embeddings and FastText embeddings. These representations were then applied to a deep learning-based CNN-BiLSTM model, which combined the strengths of Convolutional Neural Networks (CNN) for feature extraction and Bidirectional Long Short-Term Memory (BiLSTM) networks for understanding contextual information from both directions in a sequence. Our results indicated that the TF-IDF representation achieved the best performance in classifying the classes, with an AUC of 84%, which is better than both BERT and FastText embeddings. This finding highlights the effectiveness of TF-IDF in capturing the essential features of Bangla text in our dataset. Additionally, while BERT did not outperform TF-IDF in terms of AUC, it achieved a good F1 score, demonstrating its capability to effectively balance precision and recall. When comparing our work with other published research, we found that our results were superior in terms of various evaluation metrics, including accuracy, precision, recall and F1-score. Additionally, the reliability of our dataset annotations, performed by domain experts, contributed to the acceptance of our model.

**Acknowledgement**

This research is financially supported by the Xiamen University Malaysia (Project codes: XMUMRF/2021-C8/IECE/0025 and XMUMRF/2022-C10/IECE/0043).

**References**

[1] Nancy Frasure-Smith, François Lespérance and Mansour Talajic, "The impact of negative emotions on prognosis following myocardial infarction: Is it more than depression?", *Health Psychology*, vol. 14, no. 5, pp. 388–398, September 1995, Published by American Psychological Association, Print ISSN: 0278-6133, Online ISSN: 1930-7810, DOI: 10.1037/0278-6133.14.5.388, Available: https://psycnet.apa.org/doi/10.1037/0278-6133.14.5.388.
[2] Kevin M. Malone, Gretchen L. Haas, John A. Sweeney and J. John Mann, "Major depression and the risk of attempted suicide", *Journal of Affective Disorders*, vol. 34, pp. 173–185, 1995, ISSN: 0165-0327, DOI: 10.1016/0165-0327(95)00015-F, Available: https://www.sciencedirect.com/science/article/abs/pii/016503279500015F.
[3] Casey T. Carr and Rebecca A. Hayes, "Social Media: Defining, Developing, and Divining", *Atlantic Journal of Communication*, vol. 23, no. 1, pp. 46–65, 2015, Print ISSN: 1545-6870, Online ISSN: 1545-6889, DOI: 10.1080/15456870.2015.972282, Available: https://www.tandfonline.com/doi/abs/10.1080/15456870.2015.972282.
[4] John A. Naslund, Arnav Bondre, John Torous and Kelly A. Aschbrenner, "Social Media and Mental Health: Benefits, Risks, and Opportunities for Research and Practice", *Journal of Technology in Behavioral Science*, vol. 5, pp. 245–257, 2020, Electronic ISSN: 2366-5963, Published by Springer Nature, DOI: 10.1007/s41347-020-00134-x, Available: https://link.springer.com/article/10.1007/s41347-020-00134-x.
[5] Prakahs M Nadkarni, Lucila Ohno-Machado and Wendy W Chapman, "Natural language processing: an introduction", *Journal of the American Medical Informatics Association*, vol. 18, no. 5, pp. 544–551, September 2011, Published by Oxford University Press, ISSN: 1067-5027, DOI: 10.1136/amiajnl-2011-000464, Available: https://academic.oup.com/jamia/article/18/5/544/829676.
[6] Avinash Madasu and Sivasankar Elango, "Efficient feature selection techniques for sentiment analysis", *Multimedia Tools and Applications*, December 2019, Published by Springer, ISSN: 1573-7721, DOI: 10.1007/s11042-019-08409-z, Available: https://link.springer.com/article/10.1007/s11042-019-08409-z.
[7] Abdul Hasib Uddin, Durjoy Bapery and Abu Shamim Mohammad Arif, "Depression Analysis of Bangla Social Media Data using Gated Recurrent Neural Network", in *Proceedings of the 2019 1st International Conference on Advances in Science, Engineering and Robotics Technology (ICASERT)*, 03-05 May 2019, Dhaka, Bangladesh, Published by IEEE, Electronic ISBN:978-1-7281-3445-1, Print on Demand(PoD) ISBN:978-1-7281-3446-8, DOI: 10.1109/icasert.2019.8934455, Available: https://ieeexplore.ieee.org/document/8934455.

[8]  Abdul Hasib Uddin, Durjoy Bapery and Abu Shamim Mohammad Arif, "Depression Analysis from Social Media Data in Bangla Language using Long Short Term Memory (LSTM) Recurrent Neural Network Technique", in *Proceedings of the 2019 International Conference on Computer, Communication, Chemical, Materials and Electronic Engineering (IC4ME2)*, 11-12 July 2019, Rajshahi, Bangladesh, Published by IEEE, Electronic ISBN:978-1-7281-3060-6, Print on Demand (PoD) ISBN:978-1-7281-3061-3, DOI: 10.1109/IC4ME247184.2019.9036528, Available: https://ieeexplore.ieee.org/abstract/document/9036528.

[9]  Shaika Chowdhury and Wasifa Chowdhury, "Performing sentiment analysis in Bangla microblog posts", in *Proceedings of the 2014 International Conference on Informatics, Electronics & Vision (ICIEV)*, 23-24 May 2014, Dhaka, Bangladesh, Published by IEEE, Electronic ISBN:978-1-4799-5180-2, Print ISBN:978-1-4799-5179-6, DOI: 10.1109/ICIEV.2014.6850712, Available: https://ieeexplore.ieee.org/document/6850712.

[10] Farzana Tasnim, Sultana Umme Habiba, Nuren Nafisa and Afsana Ahmed, "Depressive Bangla Text Detection from Social Media Post Using Different Data Mining Techniques", *Lecture Notes in Electrical Engineering*, vol. 834, pp. 237–247, 3 March 2022, Published by Springer, Print ISBN: 978-981-16-8483-8, Online ISBN: 978-981-16-8484-5, DOI: 10.1007/978-981-16-8484-5_21, Available: https://link.springer.com/chapter/10.1007/978-981-16-8484-5_21.

[11] Abdhullah-Al-Mamun and Shahin Akhter, "Social media bullying detection using machine learning on Bangla text", in *Proceedings of the 2018 10th International Conference on Electrical and Computer Engineering (ICECE)*, 20-22 December 2018, Dhaka, Bangladesh, Published by IEEE, Print on Demand (PoD) ISBN: 978-1-5386-7483-3, Electronic ISBN: 978-1-5386-7482-6, DOI: 10.1109/icece.2018.8636797, Available: https://ieeexplore.ieee.org/document/8636797.

[12] Asif Hassan, Mohammad Rashedul Amin, Abul Kalam Al Azad and Nabeel Mohammed, "Sentiment analysis on bangla and romanized bangla text using deep recurrent models", in *Proceedings of the 2016 International Workshop on Computational Intelligence (IWCI)*, 12-13 December 2016, Dhaka, Bangladesh, Published by IEEE, Print on Demand (PoD) ISBN: 978-1-5090-5770-2, Electronic ISBN: 978-1-5090-5769-6, DOI: 10.1109/iwci.2016.7860338, Available: https://ieeexplore.ieee.org/document/7860338.

[13] Zewen Li, Fan Liu, Wenjie Yang, Shouheng Peng and Jun Zhou, "A Survey of Convolutional Neural Networks: Analysis, Applications, and Prospects", *IEEE Transactions on Neural Networks and Learning Systems*, vol. 33, no. 12, pp. 1–21, 2021, Published by IEEE, Print ISSN: 2162-237X, Electronic ISSN: 2162-2388, DOI: 10.1109/tnnls.2021.3084827, Available: https://ieeexplore.ieee.org/document/9451544.

[14] Sepp Hochreiter and Jürgen Schmidhuber, "Long Short-Term Memory", *Neural Computation*, vol. 9, no. 8, pp. 1735–1780, November 1997, Published by MIT Press, Print ISSN: 0899-7667, Online ISSN: 1530-888X, DOI: 10.1162/neco.1997.9.8.1735, Available: https://direct.mit.edu/neco/article-abstract/9/8/1735/6109/Long-Short-Term-Memory.

[15] Kazuyuki Hara, Daisuke Saito and Hayaru Shouno, "Analysis of function of rectified linear unit used in deep learning", in *Proceedings of the 2015 International Joint Conference on Neural Networks (IJCNN)*, 12-17 July 2015, Killarney, Ireland, Published by IEEE, Electronic ISSN: 2161-4407, Print ISSN: 2161-4393, DOI: 10.1109/IJCNN.2015.7280578, Available: https://ieeexplore.ieee.org/abstract/document/7280578.

[16] Xiaoyong Yuan, Zheng Feng, Matthew Norton, Xiaolin Li, "Generalized Batch Normalization: Towards Accelerating Deep Neural Networks", in *Proceedings of the AAAI Conference on Artificial Intelligence*, vol. 33, no. 01, pp. 1682-1689, 17th July 2019, Published by AAAI Press, Print ISSN: 2159-5399, Online ISSN: 2374-3468, DOI: https://doi.org/10.1609/aaai.v33i01.33011682, Available: https://ojs.aaai.org/index.php/AAAI/article/view/3985.

[17] Afia Zafar, Muhammad Aamir, Nazri Mohd Nawi, Ali Arshad, Saman Riaz et al., "A Comparison of Pooling Methods for Convolutional Neural Networks", *Applied Sciences*, vol. 12, no. 17, 2022, Published by MDPI, ISSN: 2076-3417, DOI: 10.3390/app12178643, Available: https://www.mdpi.com/2076-3417/12/17/8643.

[18] Sungheon Park and Nojun Kwak, "Analysis on the Dropout Effect in Convolutional Neural Networks", *Computer Vision -- ACCV 2016*, pp. 189–204, 10th March 2017, Published by Springer, Print ISBN: 978-3-319-54183-9, Online ISBN: 978-3-319-54184-6, DOI: https://doi.org/10.1007/978-3-319-54184-6_12, Available: https://link.springer.com/chapter/10.1007/978-3-319-54184-6_12.

[19] Imran Khan Mohd Jais, Amelia Ritahani Ismail and Syed Qamrun Nisa, "Adam Optimization Algorithm for Wide and Deep Neural Network", *Knowledge Engineering and Data Science (KEDS)*, vol. 2, no. 1, pp. 41–46, 2019, Published by Universitas Negeri Malang, Print ISSN: 2597-4602, Online ISSN: 2597-4637, DOI: 10.17977/um018v2i12019p41-46, Available: https://journal2.um.ac.id/index.php/keds/article/view/6775.

[20] Anthony J. Bowers and Xiaodong Zhou, "Receiver Operating Characteristic (ROC) Area Under the Curve (AUC): A Diagnostic Measure for Evaluating the Accuracy of Predictors of Education Outcomes", *Journal of Education for Students Placed at Risk (JESPAR)*, vol. 24, no. 1, pp. 20–46, 2019, Published by Routledge, ISSN: 1082-4669, DOI: 10.1080/10824669.2018.1523734, Available: https://www.tandfonline.com/doi/full/10.1080/10824669.2018.1523734.